# Learning to Guide: Guidance Law Based on Deep Meta-learning and Model Predictive Path Integral Control

**CHEN LIANG[1], WEIHONG WANG[1], ZHENGHUA LIU[1], CHAO LAI[2], AND BENCHUN ZHOU[1]**

[1]School of Automation Science and Electrical Engineering, Beihang University, Beijing 100191, China

[2]Navigation and Control Technology Research Institute of China North Industries Group Corporation, Beijing 100089, China

Corresponding author: Chen Liang (e-mail: tccliangchen@hotmail.com).

This work was supported in part by the National Defense Pre-Research Foundation of China under Grant B212013XXXX.

**ABSTRACT** In this paper, we present a novel guidance scheme based on model-based deep reinforcement learning (RL) technique. With model-based deep RL method, a deep neural network is trained as a predictive model of guidance dynamics which is incorporated into a model predictive path integral (MPPI) control framework. However the traditional MPPI framework assumes the actual environment similar to the training dataset for the deep neural network which is impractical in practice with different maneuvering of target, other perturbations and actuator failures. To address this problem, our method utilize meta-learning technique to make the deep neural dynamics model adapt to such changes online. With this approach we can alleviate the performance deterioration of standard MPPI control caused by the difference between actual environment and training data. Then, a novel guidance law for a varying velocity interceptor intercepting maneuvering target with desired terminal impact angle under actuator failure is constructed based on aforementioned techniques. Simulation and experiment results under different cases show the effectiveness and robustness of the proposed guidance law in achieving successful interceptions of maneuvering target.

## I. INTRODUCTION

With development of missile and tighter performance requirements of engagement, modern missiles are expected to obtain not only a small miss distance but also a desired terminal angle under complicated requirement such as target maneuver, varying velocity and actuator failures to enhance overall interception performance. These requirements elevated the difficulties in designing missile guidance systems under model mismatch, and recent advances in deep RL provides a new perspective to tackle such problem.

Missile guidance system steers missile to target by generating acceleration command to guide missile using seeker measurement and is one of the key component of missile [1]. The guidance problem is normally considered as a control problem solved by traditional control method like sliding mode tracking control [2], [3] and optimal control. As deep neural networks shows great potential [4], [5] in recent years, deep RL provides new insight into the guidance law design and is incorporated in our work.

Many control problems can be formulated as RL problems, and RL has been proven to be successful in various such tasks. In RL, the system seeks to accomplish some tasks by using data collected by itself and good results are achieved in continuous and high dimensional state space. The methods of such technique can be divided into model-free and model-based methods. Model-free methods have been applied successfully in many control problems such as quadrotor [6], multi-joint robots [7], spacecraft guidance [8], and so on. They do not need a model as system models are becoming more complex but require many interactions with the environment. The model-based deep RL methods are generally considered data efficient [9] in controlling task compared to task-specific model-free method since model-based methods usually need only dozens of minutes of experience but they generally need to construct a model to predict system dynamics. In this work, we presented a model-based RL guidance method in which a deep neural network dynamic trained using sample data is integrated into MPPI, a model predictive control (MPC) architecture. MPPI is one of the efficient method to solve the Hamilton-Jacobi-Bellman (HJB) equation by using Monte-Carlo sampling and is thus used to construct our guidance law.

Actuator failure is another important problem encountered in control of complex systems. Most methods treats the actuator failure and other external perturbation as disturbance of the system with the reference system model unchanged, which limits its performance. Meta-learning provides learning ability of the deep neural network and provides a new perspective on how to solve this problem. Thus these situations can be solved as constantly learning the changes of environment with the deep neural dynamic model online to deal with these changes via adaption. In this work, meta-learning is integrated into the MPPI control framework and provides online adaption of the system through deep neural network constant adapting to current environment in MPPI. Impact angle constraint is also important in guidance law design since it greatly improves the penetration probability in the weak point of target. In our work, the line-of-sight angles are chosen as the impact angle constraint, and the proposed guidance scheme ensure the interceptor intercept the target with desired terminal impact angles.

The main contribution of our work is as follows: 1) With model-based deep RL, the missile guidance problem can be solved in a data-driven way, without intensive mathematical modeling of guidance system in previous works using traditional control theory approach. 2) A meta-learning MPPI control method with adaptive temperature coefficient is proposed, in which the MPPI controller can adapt to changes and perturbations online through meta-learning technique. 3) A novel guidance scheme is formulated with aforementioned techniques to achieve a guidance law with desired terminal impact angle with varying velocity interceptor intercepting maneuvering target under actuator failures. To the best of our knowledge, we believe this work is the first to achieve a high-performance missile guidance law under constraint build upon recent advances in deep learning.

This paper is organized as follows. Section II reviews existing works on model-based RL, MPC guidance laws, MPPI and meta-learning. A new guidance scheme based on model-based RL and meta-learning is explained in section III. Section IV details the experiments and simulations to show the effectiveness of the proposed method. Finally, conclusions are given in section V.

## II. RELATED WORK

### A. MODEL-BASED RL

Although model-free RL achieves excellent performance in different tasks such as quadrotor control [6], multi-joint robots [7] and so on, model-based methods is welcomed in many real applications for its high data efficiency [9], since they solve the RL problem by learning the dynamic model to maximize the expected return.

Model Based learning methods have utilized various models to predict the behavior of system. Simple function approximator like time-varying linear models [10], have been used in many model-based control problems, as well as Gaussian processes [11], [12] and mixture of Gaussians. These methods achieve high sample efficiency, however they have difficulties when dealing with high-dimensional sampling spaces and nonlinear model dynamics [13].

With the limitations of the traditional models, data-intensive methods like deep learning adopt self-supervision in data collection, thus allows creation of large dataset and achieves better outcomes [14], and recent works utilizes deep feed-forward neural network or deep Long-Short-Term-Memory (LSTM) neural network for several control problems like quadrotor [15], robot-assisted dressing [16] and under-actuated legged millirobots [17]. These works achieves better outcomes than the traditional models of model-based RL.

### B. GUIDANCE LAWS WITH MPC

Extensive works have been done on guidance laws using sliding mode control [18], dynamics surface control [19] and other traditional control methods. With the development of Stochastic optimal control (SOC), MPC has been proposed in many guidance problems for its satisfactory performance and ability in providing many constraints and combining objectives. For example, [20] utilizes MPC with optimal sliding mode in guidance and [21] incorporates neural networks as optimization method in MPC guidance law. These MPC guidance laws shows high performance in guidance and MPC is one of the most effective ways to achieve generalization for RL tasks [22], thus MPC based guidance is carried out with the model-based deep RL framework in this work.

### C. MPPI

In SOC, the uncertainty in control and resultant state uncertainty are dealt with. The SOC problem usually needs the HJB equation to be solved using standard optimal control theory. In MPPI, an information theoretic SOC method, a path integral method is used to find a sequence of control commands by minimizing the running cost which is the integral of each individual cost in each step where solution to the HJB equation is approximated using importance sampling of these paths [23] by using Feynman-Kac theorem and KL divergence. As this method is easy to be implemented in parallel [24] and can cope with states costs, this methods has been popular to be integrated in the model-based RL framework with tasks like aggressive driving [25] and complex robot manipulation task [14]. Thus MPPI framework is utilized in this work, which solves the low sampling efficiency problem of simple random shooting MPC method in many model-based RL and, as an effective and flexible control framework, to formulate a guidance scheme by integrating with meta-learning.

### D. META-LEARNING

Standard model-based and model-free RL methods normally train a model or policy to be used at test time in advance based on assumption that actual environment match what they saw at training, however these is not the case and the change sometimes cause the controller to fail. When dealing with such changes like actuator failure and other perturbations in guidance problems, current guidance law normally consider such change as disturbance [26] or adjust guidance law parameters online [27]. These methods, however, limits its performance for not changing the nominal system. These limitations can be tackled by meta-learning, a long-history machine learning problem which enabling agents learn new tasks or adapts to the change by learning to learn [28], [29] and is usually done by deciding an update rule for the learner [30]. Many works develops meta-learning in model-free RL [31] [32], and recent work [33] utilizes meta-learning in model-based RL. Meta-learning have also been used in real applications such as recommendation [34], selecting [35], navigation [36] and so on [37]. Our work differs the meta-learning MPPI method in [33] with the adaptive temperature coefficient in the MPPI controller which provides a better control performance while constantly adapting to changes in environment.

## III. METHOD

In our approach, a meta-learning framework integrated a MPPI as action selection controller is set up to have the ability to adapt online and deal with our guidance problem. This framework is proposed to have better data efficiency and to adapt to changes in environment at test time, since the model-based deep RL use less experience to train and meta-learning can achieves fast adaption. Our meta-learning approach can be divided into two phases, first a meta-training phase, and then an online adaptive control phase. The neural network we use and these two phases as well as the MPPI controller in the online adaptive control phase will be described in detail below and the control method is listed in Alg. 1.

### A. NEURAL NETWORK DYNAMICS MODEL

The system dynamic model used in this meta-learning model-based deep RL framework is built as a deep neural network. Using deep learning, such a network can be learned from observations of the real system. The network can be parameterized as $f_\theta(\boldsymbol{x_t}, \boldsymbol{u_t})$, where the parameter $\theta$ is the weight coefficients in the neural network. This dynamic function takes the current states $x_t$ and current controls $u_t$ as inputs.

A fully connected neural network is set up as the specific deep neural network model. This comparably simple network is chosen to give better performance when given fewer amounts of data, since sampling is relatively hard in guidance problems and costs less time to execute in practice than other more complicated networks like LSTM.

Like in many deep neural dynamic models of deep model-based RL [17], [25], [38], the neural dynamic model predicts the derivate of the system states. In this work the neural dynamic prediction is augmented into the system states as an extended state, thus the neural dynamic model can also be seen as an extended state observer, which can be illustrated as follows:

$$\boldsymbol{\hat{x}_{t+1}} = f_\theta(\boldsymbol{x_t}, \boldsymbol{u_t}) = \begin{bmatrix} q_{t+1} \\ \dot{q}_{t+1} \\ \ddot{q}_{t+1} \end{bmatrix} = \begin{bmatrix} q_t + \dot{q}_t \Delta t \\ \dot{q}_t + \ddot{q}_t \Delta t \\ f_\theta(x_t, u_t)\Delta t \end{bmatrix}, \qquad (1)$$

where $\boldsymbol{\hat{x}_{t+1}}$ is the predicted state. The state $\boldsymbol{x}$ is partitioned into the observable state $(q, \dot{q})$ and the extended state $\ddot{q}$ , and $\Delta t$ is the discrete time increment.

### B. META-LEARNING

The neural dynamic model described above is utilized in the meta-learning framework. In this online adaption meta-learning method we adopted from [33], a two phase training is implemented to make the model optimized to the training dataset and also optimized for adaption online. In the first phase, the meta-training phase, the neural network model is trained using the dataset sampled from the system. And in the second phase, combined with recent data, the neural network can adapt quickly to the local context as shown in Fig. 1 below. As the environment is constantly evolving, we assume at every time step, the task is slight different as the changing environment the controller facing. The description of these two phases is reviewed in the rest of this section:

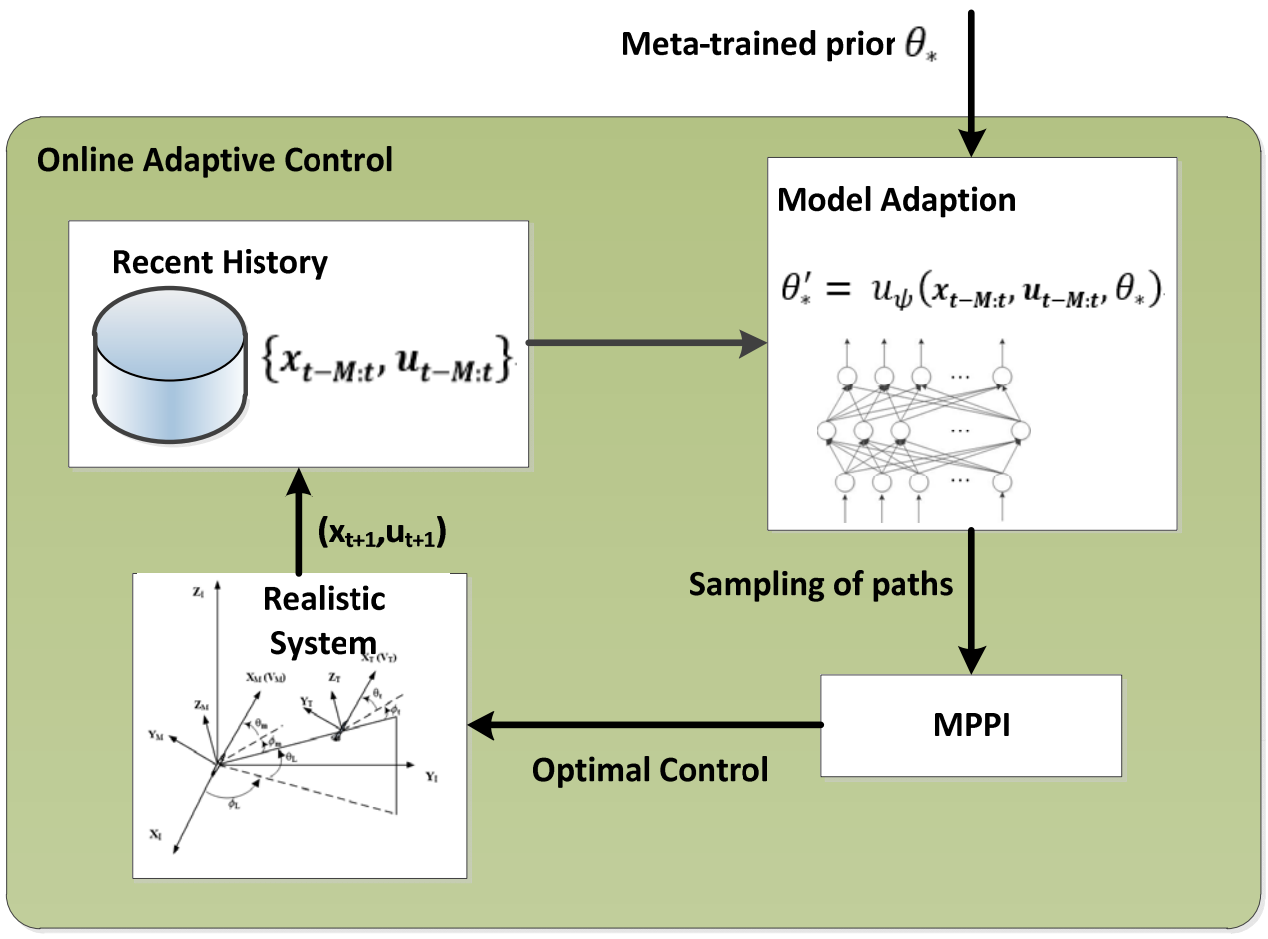

FIGURE 1. **High-level diagram of the proposed approach**

#### 1) META-TRAINING

In our framework, a meta-training step is first executed to learn an optimized neural dynamic model parameter $\theta_*$ to be further adapted online. This step is similar to normal training of a neural dynamic model during which we first collect a training dataset from the system, then the network is optimized through stochastic gradient descent.

The training dataset is collected by collecting training data starts in a random starting state and random controls with zero mean is given in each timestamp. Each individual training trajectory is recoded till terminal conditions such as missile hits the target in this case, otherwise till a predefined max timestamp. After the training data is collected, the dataset is preprocessed. Firstly the data is cut into slices consists of current state $\boldsymbol{x_t}$, control $\boldsymbol{u_t}$ and the extended state $\ddot{q}_{t+1}$ is generated through differentiate the observed $\dot{q}$. Then the data is normalized into a zero-mean and a reduced variance form to help the gradient flow in neural network training. The reason why standard deviation of some states is only reduced but not to one is for it makes the neural difficult to train in the low value portion of high variance raw data. After the data is normalized, white noise is added to the state as data augmentation and also enhance prediction performance under noise.

The model $f_\theta(\boldsymbol{x_t}, \boldsymbol{u_t})$ is trained using the training dataset by minimizing the distance between the predicted output and actual training data. Mean absolute error (MAE) is chosen instead of mean square error (MSE) to be this distance since we want to have a same scale of the high variance data. Adam optimizer [39], a stochastic gradient descent optimization method is employed to tackle this optimization problem.

### 2) ONLINE MODEL ADAPTIVE CONTROL

Given a meta-trained model $f_{\theta_*}(\boldsymbol{x_t}, \boldsymbol{u_t})$ , in the online adaption phase we want to utilize recent experience to get a more accurate predictor. Thus an update rule $u_\psi$ can be utilized to adapt the model weight coefficients $\theta_*$ into $\theta_*'$ using recent experience $\tau_\varepsilon(t-M, t-1)$. We select the update rule as a gradient ascent of the likelihood of experience through MAE between model predict and ground truth:

$$\mathcal{N}_\psi(\tau(t-M, t-1), \theta) = \theta + \alpha \nabla_\theta \left(\frac{1}{M}\sum_{m=t-1}^{t-M} \left\|(\dot{q}_{m+1} - \dot{q}_m) - \hat{f}_\theta(\boldsymbol{x_m}, \boldsymbol{u_m})\right\|\right), \quad (2)$$

where $\alpha$ is the learning rate which is set fixed in advance. MAE is used here in the update rule instead of the MSE in the original method of [33]. Since in the online adaption phase, the recent experience contains a smaller size of observations than a batch of data in the meta-training phase, resulting a less accurate estimation of step, the large step size caused by the square term in MSE cause a large step size in the high-value portion in data which deteriorate the online adaption performance.

Then a MPPI is used to output a control sequence, in which only the first control in this sequence is executed, use this updated model which is detailed later.

## C. MODEL PREDICTIVE PATH INTEGRAL CONTROLLER

Based on the dynamic model $f_{\theta_*}(x_t, u_t)$ trained above, a model-based controller can be built given a predefined cost function that describes the motivation of the control problem. The model predictive path integral control (MPPI) is a stochastic optimal control method that solves the nonlinear HJB equation [37]. In this method, a Monte-Carlo based importance sampling of multiple trajectories is used to tackle the optimal control problem. The basic framework of MPPI is from [25], and it is briefly reviewed in the rest of this section.

Consider a stochastic discrete time dynamic system in the following form:

$$\boldsymbol{x_{t+1}} = f(\boldsymbol{x_t}, \boldsymbol{u_t} + \delta\boldsymbol{u_t}), \quad (3)$$

$$\delta\boldsymbol{u_t} \sim N(0, \Sigma), \quad (4)$$

where $\boldsymbol{x_t} \in \mathbb{R}^n$ denote the state vector at discrete time t, $\boldsymbol{u_t} \in \mathbb{R}^m$ denotes the control input for this system and $\delta\boldsymbol{u_t} \in \mathbb{R}^m$ is a Gaussian noise vector. The Gaussian noise assumption is reasonable for the control inputs are normally given to a lower level servomotor controller to execute in actual system. The optimal control problem then is to find an optimal control sequence that minimize the expected value of state-dependent trajectory cost functions which has the following form:

$$S(\mathcal{E}_t^n) = \phi(\boldsymbol{x_T}) + \sum_{t=0}^{T-1} c_t(\boldsymbol{x_t}, \boldsymbol{u_t}), \quad (5)$$

where the $c_t$ and $\phi$ are respectively running and terminal cost function and $\mathcal{E}_t^n$ denote the noise sequence.

Define $V$ to be a sequence of inputs to this system. Then three distinct distributions for $V$ can be probability density function of an input sequence in the uncontrolled system, an open-loop control sequence and the optimal control sequence of this problem, which can define to be $\boldsymbol{h}$ , $\mathbf{p}$ and $\boldsymbol{p}^*$respectively. Then the free energy of the dynamic system is defined to be:

$$\mathcal{F}(V) = -\lambda \log(\mathbb{E}_\mathbb{P}[\exp(-\frac{1}{\lambda}S(V))]), \quad (6)$$

where $\lambda$ is a positive scalar. Since the cost of optimal control problem is bounded from below from this free energy [40], thus by apply Jensen's inequality and further derivation, the optimal distribution for which such bound is tight can be defined to be:

$$\boldsymbol{p}^*(V) = \frac{1}{\eta}\exp\left(-\frac{1}{\lambda}S(V)\right)\boldsymbol{h}(V), \quad (7)$$

where $\eta$ is the normalizing factor. Then the optimal control problem corresponds to minimize the gap between the controlled distribution and the optimal control distribution measured by KL divergence:

$$U^*(V) = \underset{U}{\operatorname{argmin}}\, \mathbb{D}_{KL}(\mathbb{P}^* || \mathbb{P}). \quad (8)$$

Then after expanding out the KL divergence, and analyzing the resulting concave function, the optimal control sequence can be derived as follows:

$$\boldsymbol{u}_t^* = \int \boldsymbol{p}^*(V)\boldsymbol{v_t}dV. \quad (9)$$

Since we cannot directly sample from the optimal distribution $\mathbb{Q}^*$, importance sampling technique is adopted to get the expected value of the optimal control sequence with respect to $\mathbf{p}$:

$$\boldsymbol{u}_t^* = \mathbb{E}_P[w(V)\boldsymbol{v_t}], \quad (10)$$

where the importance sampling weight $w(V)$ can be defined according to importance sampling technique:

$$w_t(V) = \frac{\boldsymbol{p}^*(V)}{\boldsymbol{h}(V)}\exp\left(\sum_{t=0}^{T-1}(-\boldsymbol{v}_t^T\Sigma^{-1}\boldsymbol{u_t} + \frac{1}{2}\boldsymbol{u}_t^T\Sigma^{-1}\boldsymbol{u_t})\right), \quad (11)$$

$$= \frac{1}{\eta} exp\left(-\frac{1}{\lambda}\left(S(U+\mathcal{E}) + \lambda\sum_{t=0}^{T-1}\frac{1}{2}\boldsymbol{u}_t^T\Sigma^{-1}(\boldsymbol{u_t} + 2\delta\boldsymbol{u_t})\right)\right). \quad (12)$$

**Remark 1.** The $\lambda$ here can be seen as the temperature coefficient in this softmax distribution. In the standard MPPI, $\lambda$ is chosen to be a predefined hyper-parameter [25]. However since the temperature $\lambda$ differs dominant index's impact in the softmax distribution, as the data variance varies in different parts of state of the system, this fixed setting will vary the performance of output control sequence. Thus the temperature $\lambda$ in our approach is computed in the following adaptive way:

$$\lambda = \lambda^{*}\sigma\big(S(V)\big), \tag{13}$$

where $\sigma$ denote the standard deviation function and $\lambda^{*}$ is a predefined temperature coefficient. This adaptive temperature can also be interpreted as the costs $S(V)$ is firstly normalized into standard deviation one which is a common data preprocessing technique. This aids the importance sampling by ensuring the weights distribution property equal in different parts of the state.

Then the MPPI update the control sequence can be seen as iterative update law from N samples as:

$$\boldsymbol{u}_t^{i+1} = \boldsymbol{u}_t^{i} + \sum_{n=1}^{N} w_t^n \delta \boldsymbol{u}_t^n. \tag{14}$$

Thus the control command can be seen as an expectation over the sample trajectories. By combining online adaption of the meta-learning neural dynamics, our proposed meta-learning MPPI control with adaptive temperature can be given in Alg. 1.

**Algorithm 1**: Online Adaptive control with MPPI at each time stamp

**Given:** $\hat{f}_\theta(\boldsymbol{x}_t, \boldsymbol{u}_t)$: transition model with parameter $\theta$ of a prior;
$\mathcal{N}_\psi$: update rule of parameter $\theta$;
$\tau(t-M, t-1)$: experience;
$N$: Number of samples;
$T$: Horizon;
$\Sigma, r_t, \phi$: Control hyper-parameters

```
θ'_* ← N_ψ(τ(t − M, t − 1), θ_*)
for n = 0, … , N − 1 do
    x ← x_0
    Sample E_t^n = {δu_0^n, δu_1^n, …, δu_{T−1}^n}
    for t = 1, … , T do
        x_t ← f̂_{θ'_*}(x_{t−1}, u_{t−1} + δu_{t−1}^n)
        S(E_t^n) += c_t(x_t, u_t) + λ u_{t−1}^T Σ^{−1} δu_{t−1}^n
    end for
    S(E_t^n) += φ(x_T)
end for
S'(E_t^n) = S(E_t^n) − min_n[S(E_t^n)]
λ = σ(S'(E_t^n))
η = Σ_{n=0}^{N−1} exp(−(1/λ)(S'(E_t^n)))
for n = 0, … , N − 1 do
    w_t^n ← (1/η) exp(−(1/λ)(S'(E_t^n)))
end for
for t = 0, … , T − 1 do
    u_t += Σ_{n=1}^{N} w_t^n δu_t^n
end for
SendControlCommand(u_0)
for t = 0, … , T − 1 do
    u_{t−1} ← u_t
end for
u_T ← Initialize(u_{T−1})
```

## IV. SIMULATION AND EXPERIMENT

The proposed meta-learning MPPI control method with adaptive temperature coefficient is tested on a guidance task and compared with standard MPPI from [25] but with our adaptive temperature method and standard meta-learning MPPI from [33] which has a fixed temperature setting. In this section, the mathematical model of three-dimensional purser-target scenario is set up, and the effectiveness of proposed control framework is validated through simulation and experiment of multiple engagements.

### A. MODEL FOR SIMULATION

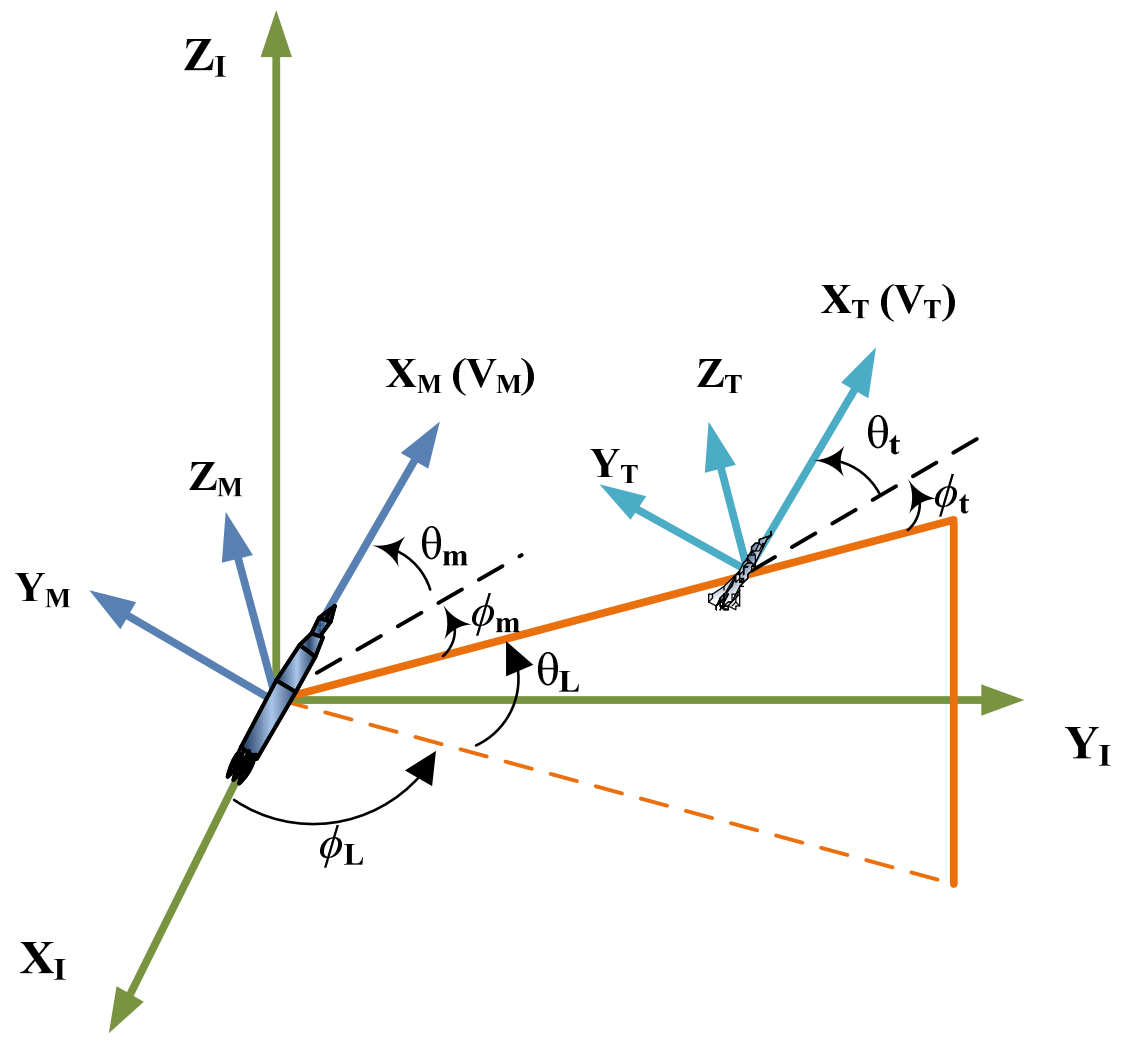

FIGURE 2. **Interception geometry used in simulation**

The three-dimensional purser-target homing guidance geometry is shown in Fig. 2. In this figure, $O_I X_I Y_I Z_I$, $O_M X_M Y_M Z_M$ and $O_T X_T Y_T Z_T$ represents the inertial, interceptor and target frames of reference, respectively. In this geometry, the interceptor has a velocity $V_M$, with direction defined by $\theta_m$ and $\phi_m$. The target has a velocity $V_T$, with direction defined by $\theta_t$ and $\phi_t$. Line-of-sight (LOS) angles are denoted by $\theta_L$ and $\phi_L$. Then the three-dimensional engagement dynamics can be expressed as follows [41]:

$$\dot{R} = V_T cos\theta_t cos\phi_t - V_M cos\theta_m cos\phi_m, \tag{15}$$

$$R\dot{\theta}_L = V_T sin\theta_t - V_M sin\theta_m, \tag{16}$$

$$R\dot{\phi}_L cos\theta_L = V_T cos\theta_t sin\phi_t - V_M cos\theta_m sin\phi_m, \tag{17}$$

$$\dot{\theta}_m = \frac{a_{zm}}{V_M} - \dot{\phi}_L sin\theta_L sin\phi_m - \dot{\theta}_L cos\phi_m, \tag{18}$$

$$\dot{\phi}_m = \frac{a_{ym}}{V_M cos\theta_m} + \dot{\phi}_L tan\theta_m cos\phi_m sin\theta_L - \dot{\theta}_L tan\theta_m sin\phi_m - \dot{\phi}_L cos\theta_L, \tag{19}$$

$$\dot{\theta}_t = \frac{a_{zt}}{V_T} - \dot{\phi}_L sin\theta_L sin\phi_t - \dot{\theta}_L cos\phi_t, \tag{20}$$

$$\dot{\phi}_t = \frac{a_{yt}}{V_T cos\theta_t} + \dot{\phi}_L tan\theta_t cos\phi_t sin\theta_L - \dot{\theta}_L tan\theta_t sin\phi_t - \dot{\phi}_L cos\theta_L. \tag{21}$$

### B. PROBLEM FORMULATION

The states is listed as $x_t = [R, \theta_m, \phi_m, \theta_L, \phi_L, \dot{R}, \dot{\theta}_m, \dot{\phi}_m, \dot{\theta}_L, \dot{\phi}_L, \dot{R}, \dot{\theta}_m, \dot{\phi}_m, \ddot{R}, \ddot{\theta}_m, \ddot{\phi}_m, \ddot{\theta}_L, \ddot{\phi}_L]^T$, where the last five are the extended state we estimate by deep neural dynamic model. The observable states are $[R, \theta_m, \phi_m, \theta_L, \phi_L, \dot{R}, \dot{\theta}_m, \dot{\phi}_m, \dot{\theta}_L, \dot{\phi}_L]^T$. The neural dynamic function $\widehat{f_\theta}(s_t, a_t)$ is implemented as a deep feed-forward neural network model that consists of two hidden layers, each of dimension 200, with ReLU activations. To do a meta-training, we collect a system identification dataset with 30 minutes of training data. The training dataset is generated using a controller that generates random command, with the missile velocity is constant at 800m/s and target velocity at 270m/s with no maneuvers.

The buildup of model error is one of the challenges in model-based RL. With normally a short prediction horizon is needed by MPC, we can see from the prediction performance shown in histograms and curves in Fig. 3 that the prediction error is relative small, thus adequate predicting performance can be expected.

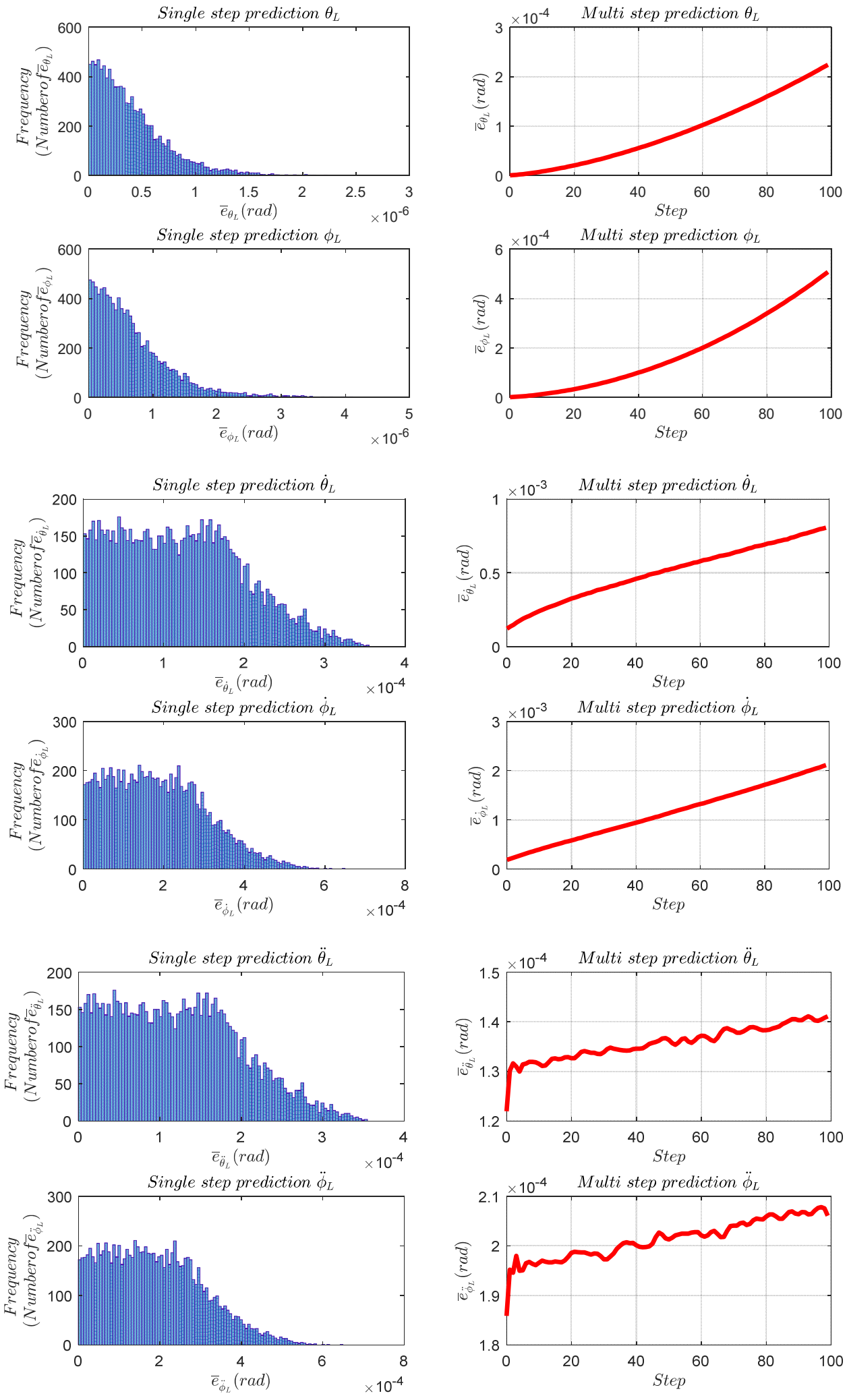

FIGURE 3. **Single-step and multi-step prediction error**

We are interested in achieving the interception of the interceptor and the target, with desired terminal angles as an additional objective, which is equivalent to nullifying the LOS angular rates $\dot{\theta}_L$ and $\dot{\phi}_L$ while controlling the LOS angles $\theta_L$ and $\phi_L$ to the desired values $\theta_{LD}$ and $\phi_{LD}$ [42]. Thus the guidance law is designed to guarantee the following equations:

$$\dot{\theta}_L = \dot{\phi}_L = 0, \quad \theta_L = \theta_{LD}, \quad \phi_L = \phi_{LD}. \tag{22}$$

**Remark 2.** Based on (22), a derivative of LOS angular rates reference planning algorithm is defined as:

$$\boldsymbol{e_0} = [\theta_L - \theta_{LD}, \phi_L - \phi_{LD}]^T, \tag{23}$$

$$\boldsymbol{e_1} = \dot{\boldsymbol{e}}_{\boldsymbol{0}} + K_1 \boldsymbol{e_0}, \tag{24}$$

where $K_1 > 0, K_2 > 0$, and the planned reference is as follows:

$$\dot{\boldsymbol{e}}_{\boldsymbol{1}} = -K_2 \boldsymbol{e_1}. \tag{25}$$

Different from normally in MPC based guidance law in which the MPC directly drive the error variables in equation (22) to zero [20], [21] via tracking, this method plan the tracking reference using similar way as a sliding surface of a sliding mode. Thus the adjustable approaching rate can make a smoother change in the error variable. If a Lyapunov function is chosen as $\mathrm{V} = \frac{1}{2}\boldsymbol{e_1}^2$, then we can derive:

$$\begin{aligned}\dot{\mathrm{V}} &= \boldsymbol{e_1}\dot{\boldsymbol{e}}_{\boldsymbol{1}} \\ &= -K_2\boldsymbol{e_1}^2 \le 0.\end{aligned} \tag{26}$$

Then we can see that the predesigned sliding surface $\boldsymbol{e_1}$ is stable subspace of the state space under the planned reference. Thus the designed error variables will converge to zero and the control requirement in (22) is satisfied.

**Remark 3:** Based on the reference planning method described above, the state dependent cost function of MPPI can be defined as follows:

$$\boldsymbol{e_2} = \dot{\boldsymbol{e}}_{\boldsymbol{1}} + K_2 \boldsymbol{e_1}, \tag{27}$$

$$\mathrm{c}(\mathbf{x}) = \|\boldsymbol{e_2}\|^2, \tag{28}$$

Thus the MPPI acts as the tracking controller to track the reference generated by Remark 2.

### C. SIMULATION RESULTS

Numerical simulations of two scenarios are performed to illustrate the performance of the proposed guidance law. The initial case parameters are given in table I and the sample and control time cycle is set to 5ms. In the simulation, a realistic interceptor velocity model from [43] is used. In this interceptor velocity model, the derivate of velocity is formulated as a function of thrust and drag which is listed below:

$$\dot{V}_M = \frac{T - D}{m} - g(\cos\phi_m \cos\theta_m \sin\theta_L + \sin\theta_m \cos\theta_L), \tag{29}$$

$$D = D_0 + D_i; \quad D_0 = C_{D0} Q s, \tag{30}$$

$$D_i = \frac{K m^2 (a_{ym}{}^2 + a_{zm}{}^2)}{Qs}, \tag{31}$$

$$K = \frac{1}{\pi A_r e}, \tag{32}$$

$$Q = \frac{1}{2}\rho V_M^2 V_M^2, \tag{33}$$

$$T = \begin{cases} 7500 & 0 \le t \le T_B, \\ 0 & t > T_B, \end{cases} \tag{34}$$

$$m = \begin{cases} 90.035 + 3.31(T_B - t) & 0 \le t \le T_B, \\ 90.035 & t > T_B, \end{cases} \tag{35}$$

where $T$, $D$, $D_0$ and $D_i$ denote thrust, drag, zero-lift drag and induced drag, and $C_{D0}$, $m$, $K$, $A_r$, $e$, $\rho$, $s$ and $Q$ are zero-lift drag coefficient, interceptor mass, induced frag coefficient, aspect ratio, efficient factor, atmosphere density, reference area and dynamic pressure which their value can be found in [43], and $T_B$ is the time of thrust exists for the interceptor.

The basic actuator failure includes lock in place, hard-over, loss of effectiveness and float [44]. We mainly concern the loss of effectiveness in this work. Commonly seen model for the loss of effectiveness is that the actuator gain $\eta$ decreases from $\eta = 1$ to a positive value less than one. And thus the actuator failure can be considered as follows [26]:

$$u_i = \begin{cases} u_{ci}\ , \quad \eta_i = 1 \quad \text{, no fault,} \\ \eta_i u_{ci}, 0 < \eta_i < 1, \text{loss of effectiveness,} \end{cases} \tag{36}$$

where $u_{ci}$ is the command signal from the $i$th controller, $u_i$ is the output of the $i$th actuator and $\eta_i$ is the $i$th actuator gain.

In order to evaluate the performance of proposed guidance law, two cases shown below with different LOS angle, interceptor and target heading are considered.

TABLE I
CASE PARAMETERS

| Dataset | Case 1 | Case 2 | Monte Carlo |
|---|---|---|---|
| R(0) (m) | 4000 | 4000 | 3800 |
| $\theta_L(0)$ (rad) | -0.7 | -0.7 | $Unif(-0.7,-0.9)$ |
| $\phi_L(0)$ (rad) | 0.65 | 0.9 | $Unif(0.5,0.7)$ |
| $\theta_m(0)$ (rad) | -0.36 | -0.1 | $Unif(-\frac{\pi}{9},\frac{\pi}{9})$ |
| $\phi_m(0)$ (rad) | -0.2 | -0.07 | $Unif(-\frac{\pi}{9},\frac{\pi}{9})$ |
| $V_M(0)$ (m/s) | 800 | 800 | 800 |
| $\theta_t(0)$ (rad) | -0.32 | 0.592 | 0.0 |
| $\phi_t(0)$ (rad) | -0.22 | -0.706 | 0.0 |
| $V_T$ (m/s) | 270 | 270 | 270 |
| $\theta_{LD}$ (rad) | -0.6 | -0.6 | -0.8 |
| $\phi_{LD}$ (rad) | 0.8 | 0.8 | 0.6 |
| $a_{yt}$ (m/s$^2$) | 40sin(t) | 40sin(t) | 30sin(t) |
| $a_{zt}$ (m/s$^2$) | 40sin(t) | 40sin(t) | 30sin(t) |
| $T_B$(s) | 3.5 | 3.0 | 4.0 |
| $\eta$ | 0.5 | 0.5 | 0.6 |
| $T_\eta$(s) | $[3.0s,+\infty)$ | $[3.0s,+\infty)$ | $[3.5s,+\infty)$ |

The initial engagement parameters in the simulation is given in table I, where $Unif$ means a uniform distribution, $\eta$ is the actuator gain described in (36), $T_\eta$ is the time period actuator failure exists and the interceptors in these cases have a max 200 m/s$^2$ acceleration capability. The MPPI controllers in our case use a horizon of 3 with control cycle of 5ms, 1000 trajectories drawn, $K_1, K_2$ in the cost function set to $[0.6, 0.5]$ and $[3.0, 2.0]$ respectively, and the temperature coefficient $\lambda^*$ is set to 1. To make a fair comparison, the temperature coefficient in standard meta-learning MPPI is set to 0.001 and 0.0001 respectively for case 1 and 2 to show the impact of different temperature setting. These values are approximately in the upper and lower range of the adaptive temperature term in the proposed method. The neural network used in the simulation is with two hidden layers of 200 neurons, ReLU activation and the learning rate $\alpha$ for online adaptive control set to 0.001.

The simulation result is shown in table Ⅱ, where the $\phi_{LT}$ and $\phi_{LT}$ represent the terminal LOS angles. We can see from the results that the proposed method achieved better result than standard MPPI from [25] but with our adaptive temperature and meta-learning MPPI method from [33] with a fixed temperature setting.

TABLE II
SIMULATION RESULT

| Case No. | Miss Distance (m) | $\theta_{LT}$ (rad) | $\phi_{LT}$ (rad) | Impact Time(s) |
|---|---|---|---|---|
| Proposed Method | | | | |
| 1 | 1.82×10$^{-5}$ | -0.6010 | 0.8039 | 8.71 |
| 2 | 1.77×10$^{-5}$ | -0.6007 | 0.8027 | 7.53 |
| Standard Meta-learning MPPI | | | | |
| 1 | 9.67×10$^{-5}$ | -0.6067 | 0.8058 | 8.88 |
| 2 | 8.77×10$^{-5}$ | -0.6120 | 0.8557 | 7.53 |
| Standard MPPI | | | | |
| 1 | 0.1007 | 0.1417 | 2.5044 | 8.76 |
| 2 | 0.2933 | 0.1842 | 2.6047 | 7.55 |

For case 1, the initial engagement parameters are selected as case 1 in table I and a comparison study of the proposed guidance law with standard MPPI and meta-learning MPPI with a fixed high temperature is carried. From table II we can see the proposed guidance law achieves better performance in miss distance, impact angle error and impact time. The simulation results, including curves of the trajectories of the interceptors and target, LOS angular rates, LOS angles, $a_{ym}$, $a_{zm}$and interceptor velocity are shown in the Fig. 5-7 below. From Fig. 4 we can see all these guidance laws can guide the interceptor to the target. The LOS angles and angular rates curves obtained by these guidance laws are shown in Fig.5. By contrast, it can be seen that the LOS angular rates obtained by standard MPPI diverge at the end of the guidance phase, resulting a larger LOS angle tracking error compared to proposed method which can show the performance of meta-learning adapting to changes in environment. We can also see the standard meta-learning MPPI have trouble tracking the desired LOS angular rate at beginning when the variance of sampled cost is small using a high temperature, since high temperature acts the importance sampling weights towards unweighted average which affect sampling sensitivity. From Fig 6, it can be seen that after the actuator partial failed, the methods with meta-learning learn the change, and adapt the neural network dynamic model accordingly, which results fast recovery of controls. We can see that the control commands of standard meta-learning MPPI saturate and have trouble keeping up in the beginning when the sampled cost has a low value variance for efficient importance sampling. Fig.7 shows the curves of the velocity of the interceptor and relative distance between interceptor and target.

We can also see from these figures that the performance of these two guidance is similar at beginning, which proves the nominal performance of standard MPPI guidance law when the environment differs not too much with training data. Consequently we can see the proposed guidance law achieves satisfactory performance with target maneuver, varying interceptor velocity and actuator failures.

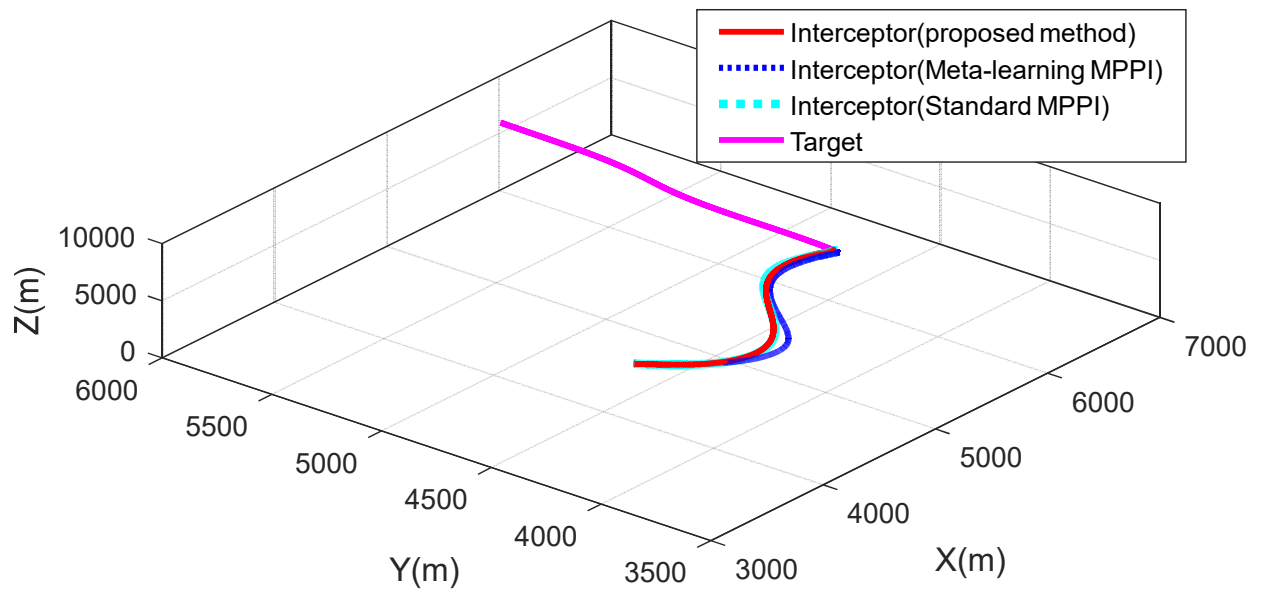

FIGURE 4. **Trajectories of interceptors and target in case 1**

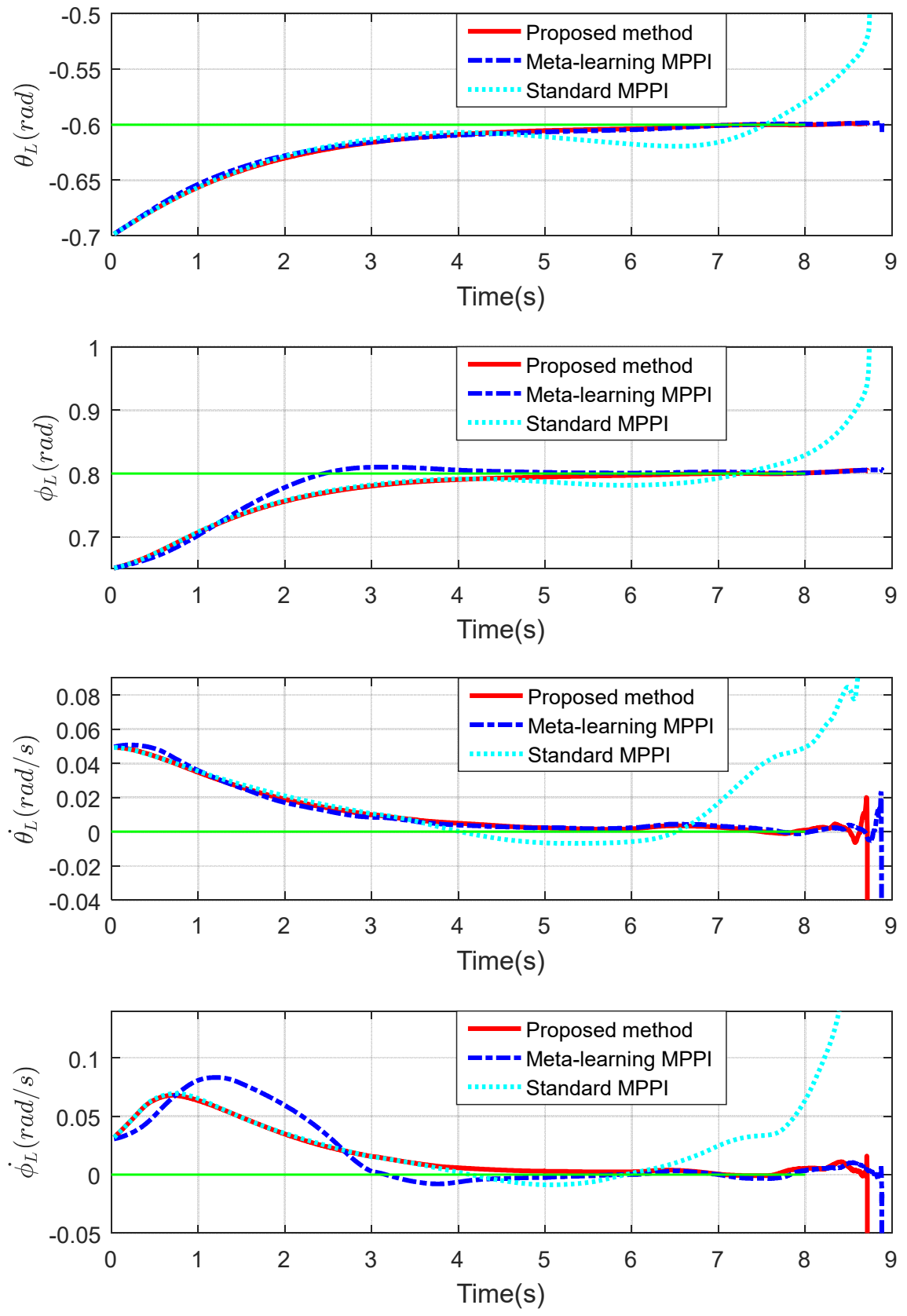

FIGURE 5. **LOS angles and angular rates in case 1**

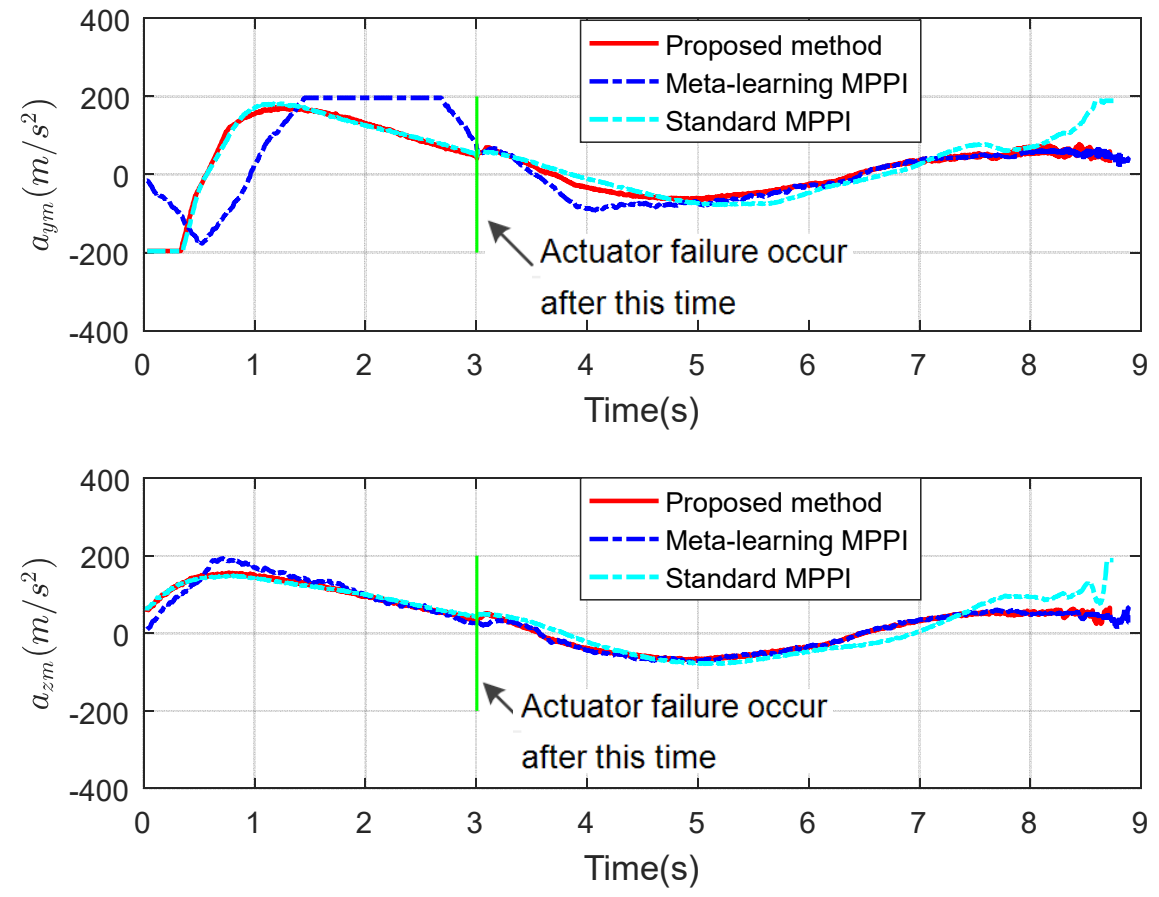

FIGURE 6. **Interceptor accelerations command in case 1**

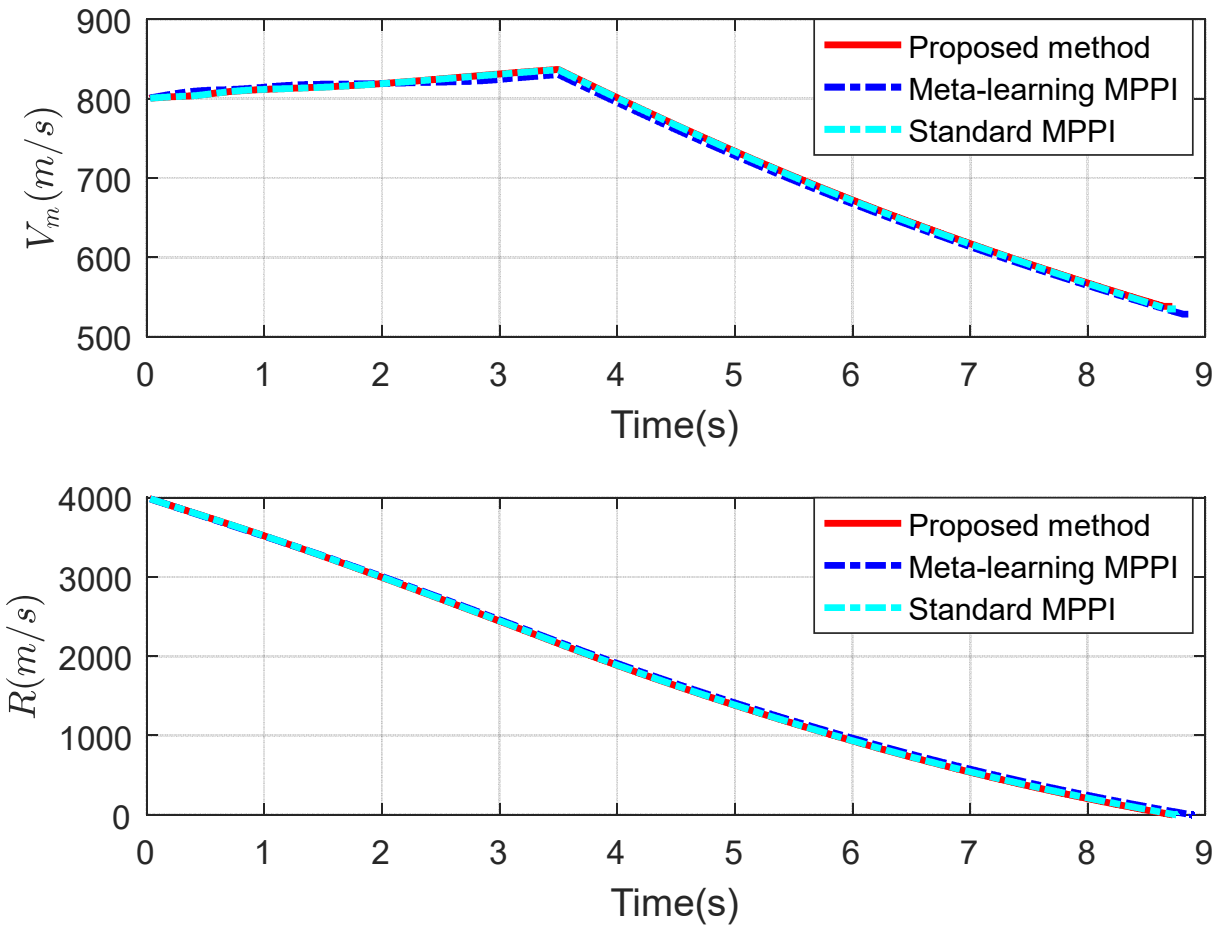

FIGURE 7. **Interceptor velocity and relative distance in case 1**

For case 2, to prove the effectiveness of the proposed guidance law, a similar simulation with case 1 but for different initial engagement parameters selected as case 2 in table I and low temperature in MPPI is presented. The simulation results, including curves of the trajectories of the interceptors and target, LOS angular rates, LOS angles, interceptor acceleration command, interceptor velocity and relative distance are shown in the Fig. 8-11. We can see from these figures the proposed guidance law achieves similar performance under a different scenario. We can see from Fig. 10 that the control command chatters in standard meta-learning MPPI at end, since the low temperature resulting many trajectories being rejected. These chatters also cause the LOS angles and angular rates to diverge. When compared with standard methods, we can see the proposed method achieves better LOS angle and angular rates tracking performance, resulting a better terminal impact angle, impact time and miss distance as shown in table II.

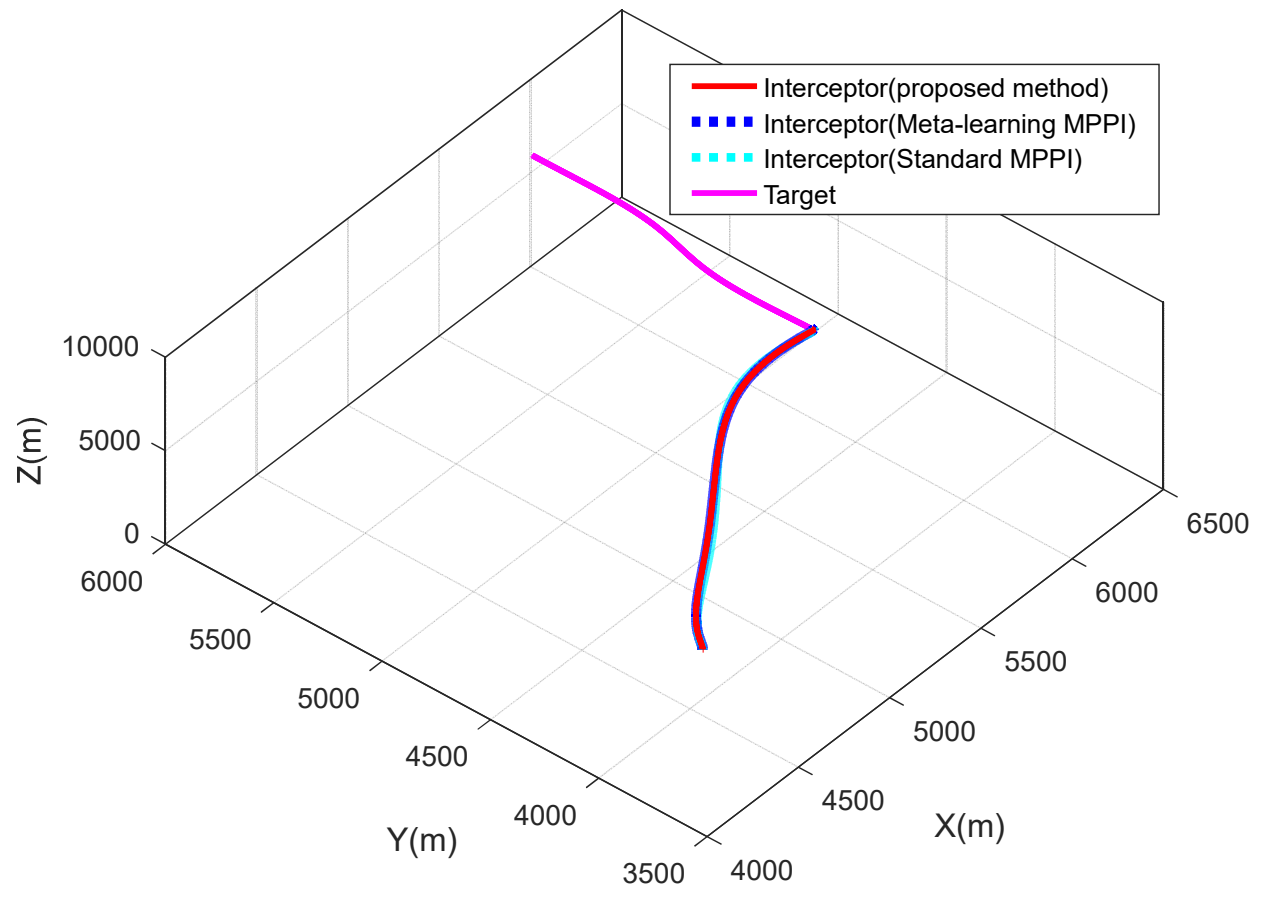

FIGURE 8. **Trajectories of interceptors and target in case 2**

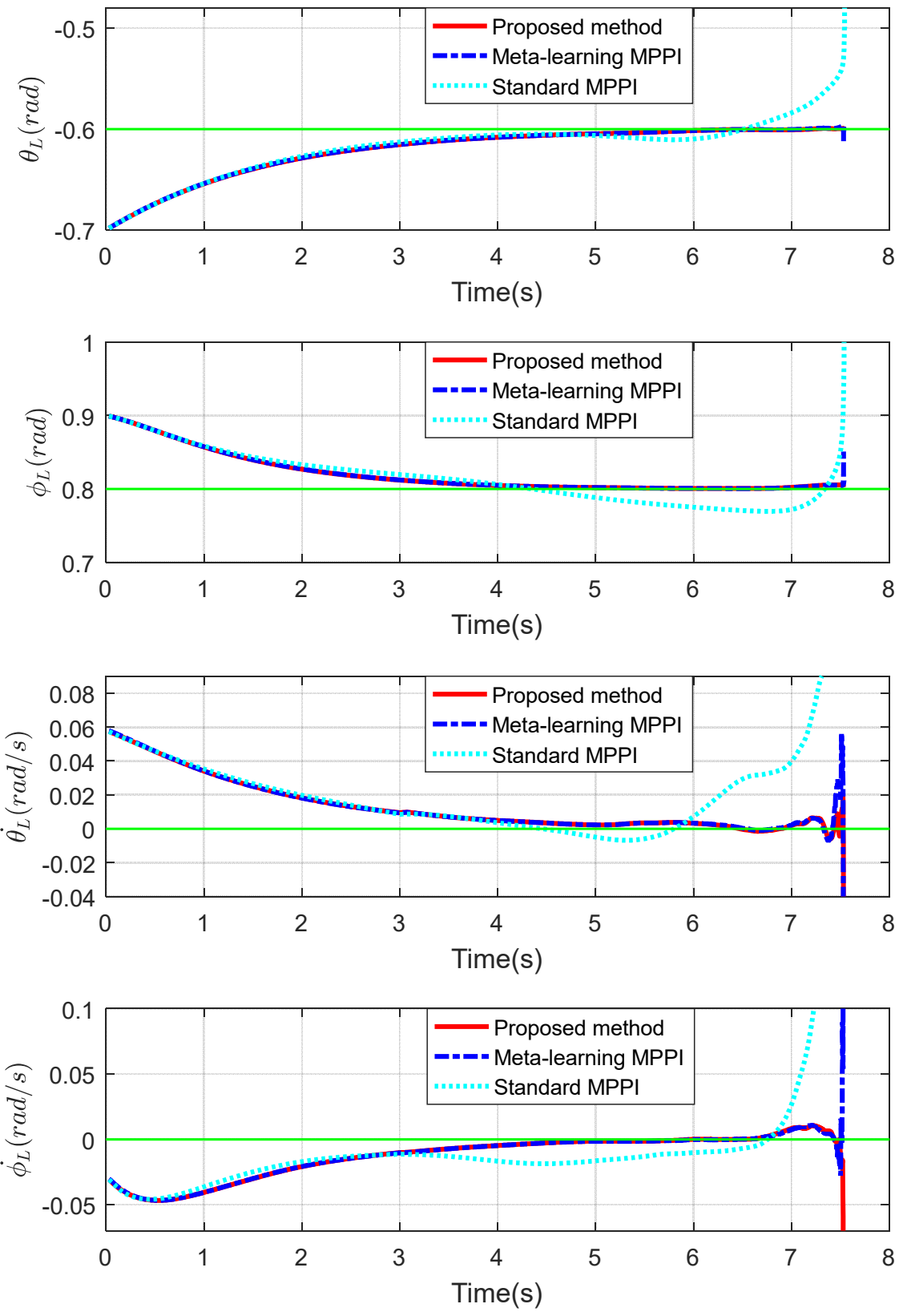

FIGURE 9. **LOS angles and angular rates in case 2**

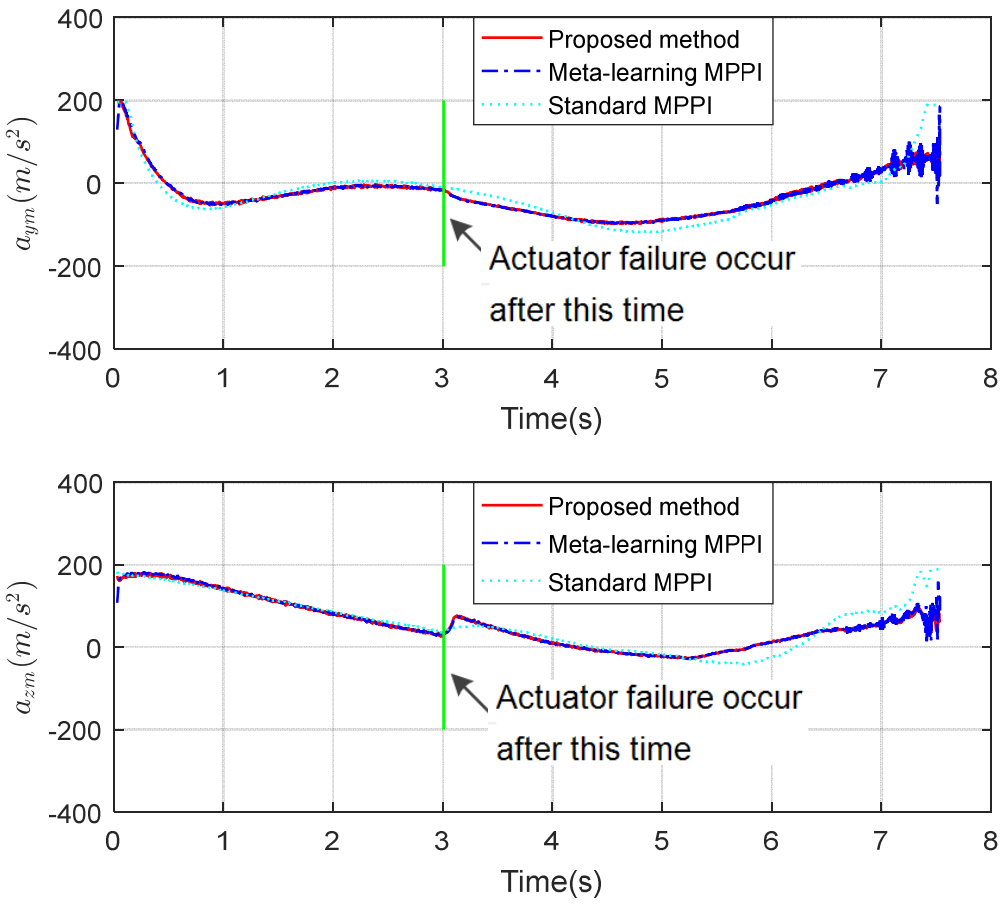

FIGURE 10. **Interceptor accelerations command in case 2**

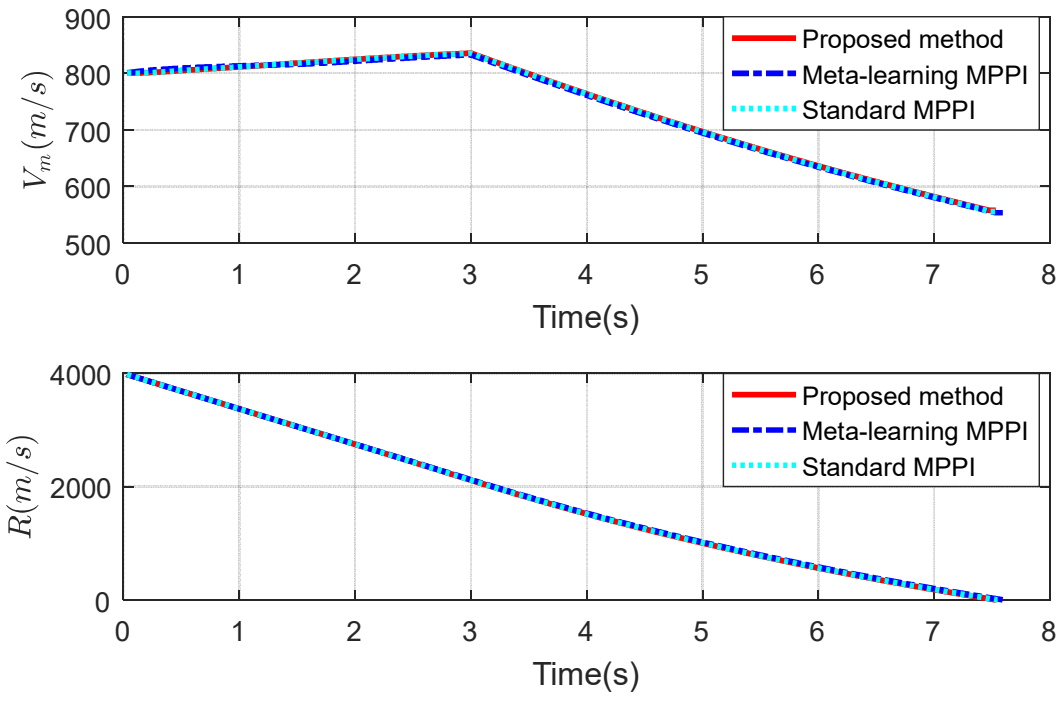

FIGURE 11. **Interceptor velocity and relative distance in case 2**

For the Monte Carlo case, to prove the effectiveness of the proposed guidance law under various initial engagement conditions, 3500 engagements are conducted with the proposed guidance law with initial conditions show in table I. The Monte Carlo sampling is carried in different initial engagement geometries which shows the performance and robustness of proposed guidance law. In this simulation, the actual observation $[R, \theta_m, \phi_m, \dot{R}, \ \dot{\theta}_m, \dot{\phi}_m]^T$ is generated by multiplying corresponding values with $1 + 0.15\sin(t)$ to simulate observation uncertainty. The measurement noise of LOS is chosen as Gaussian noise with standard deviation of 8mrad. Angular rate measurement noise of LOS is also chosen as Gaussian noise with standard deviation of one percent its current measurement value. The initial condition of the LOS angles and missile heading are set to a uniform distribution to cover the operating initial conditions, and the target heading is set to a fixed angle to make the result figures clearer to see. The actuator gain $\eta$ is set to 0.6, which means it has a 40% loss of effectiveness. All trajectories of the interceptors and target in the Monte Carlo simulation are transformed to origin at the intercept point to make the trajectories easy to see and are shown in Fig. 12. The LOS angle curves are shown in Fig.14, we can see from this figure that the LOS angles converge to the desired LOS angle. As some trajectories have initial engagement geometry that have initial LOS angular rate drive the LOS angles away from desired angles at beginning, causing them to have a more diverge LOS at around 6s till intercept. Final LOS angles, miss distances and interceptor velocities are shown in the Fig. 13 and we can see the impact angles lie around the desired angle with only minor difference and small miss distance. We can see from these figures the proposed guidance law achieves satisfactory performance under different scenarios.

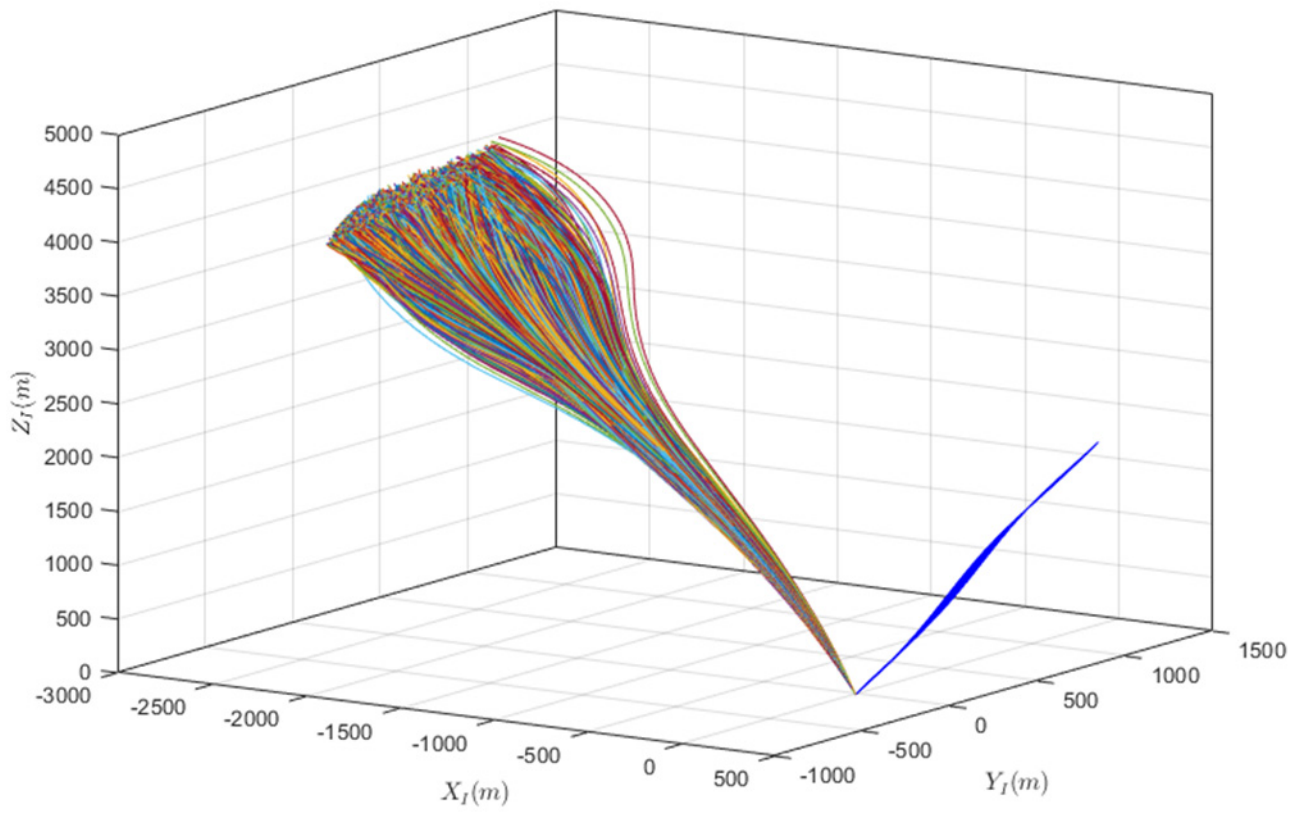

FIGURE 12. **Trajectories of interceptors (multi-color) and target (blue) in Monte Carlo case**

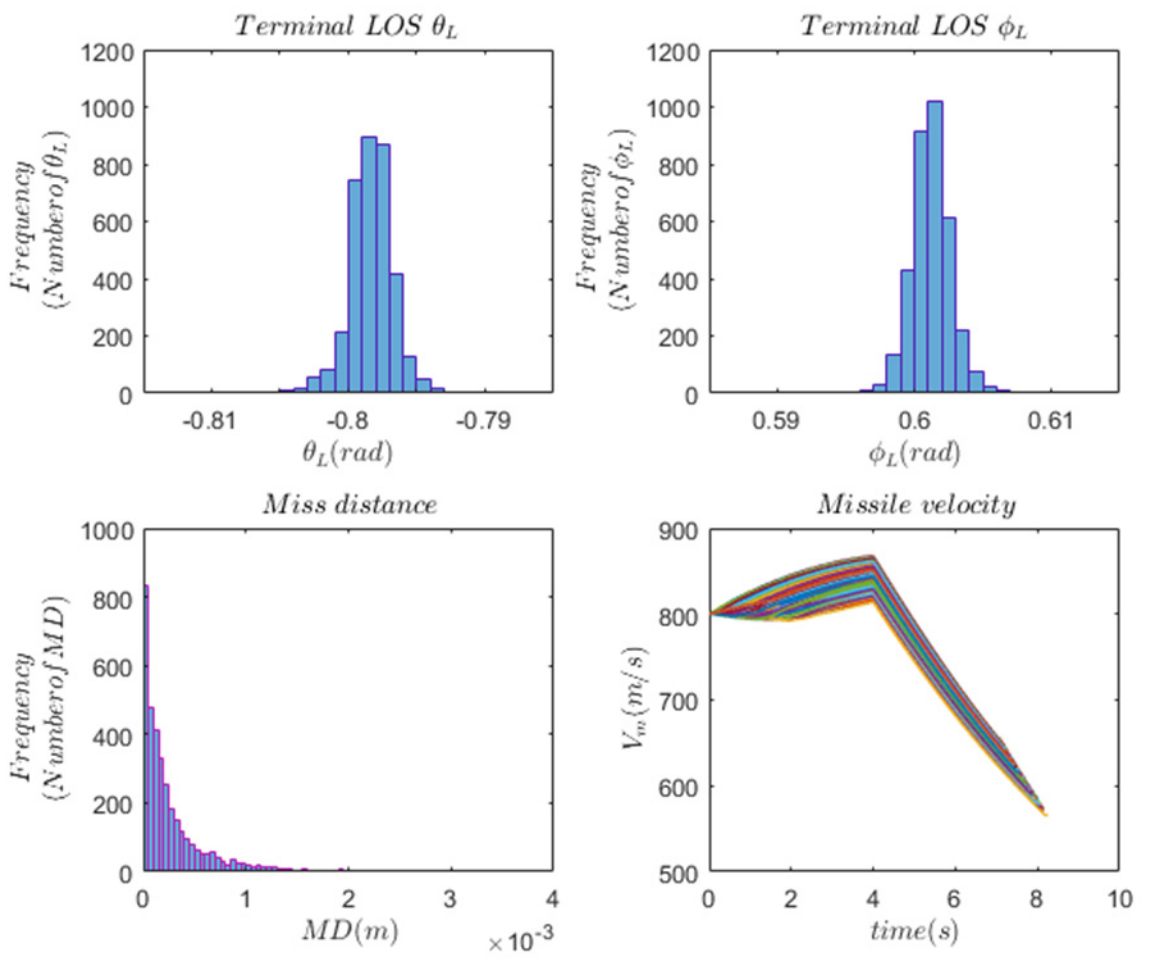

FIGURE 13. **Histogram of terminal LOS angles, miss distance, and curves of interceptor velocity in Monte Carlo case**

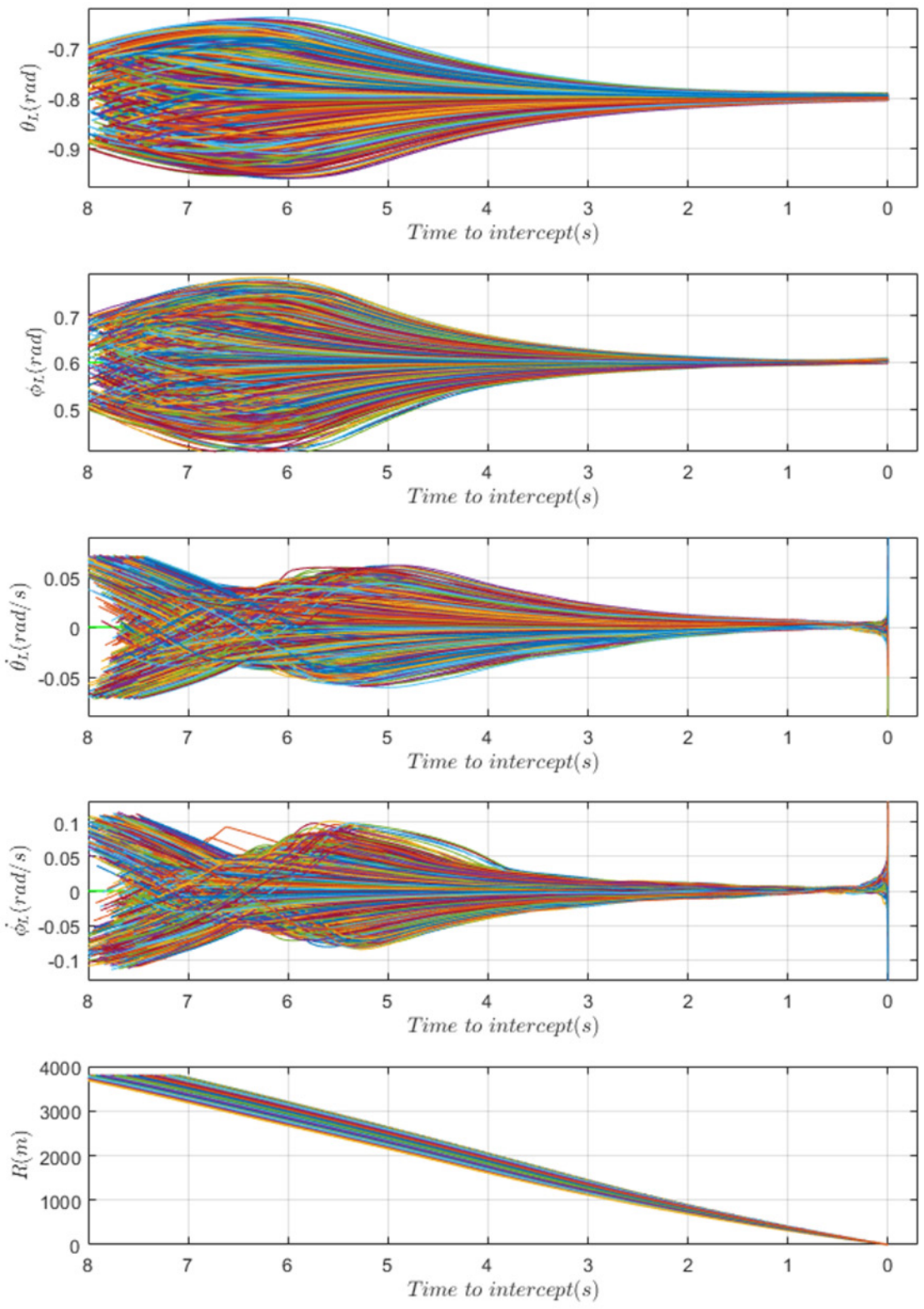

FIGURE 14. **LOS angles, angular rates and relative distance in Monte Carlo case**

## D. EXPERIMENT RESULTS

We apply our guidance law on a semi-physical simulation of a skid-to-turn missile based on a three-axis flight simulator with high precision as Fig.15 shows. In the experiment, a dynamic surface controller is constructed as the missile autopilot which receives the acceleration command from our guidance law and control the deflection of aerodynamic control surface. The missile in this experiment is axis-symmetrical with aerodynamic coefficients listed in [45], and is assumed to be roll-stabilized at a constant roll attitude. The three-axis flight simulator tracks roll, pitch and yaw attitude motion of missile in real-time and the vision guidance subsystem cooperate with the flight simulator is based on a CCD camera to generate the relative motion information, which is same as in [45]. Our guidance law generate acceleration command based on the relative motion information in the guidance computer. All subsystem communicate with the main control computer, which serve as hub, autopilot and simulator computer, through switched gigabit Ethernet. Meanwhile, the engagement is rendered in the virtual reality subsystem.

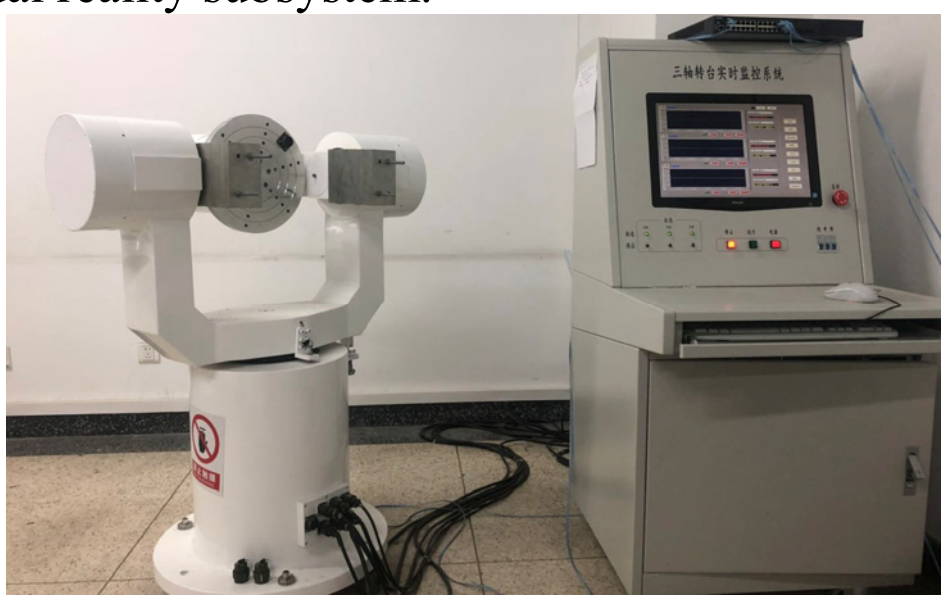

(a)

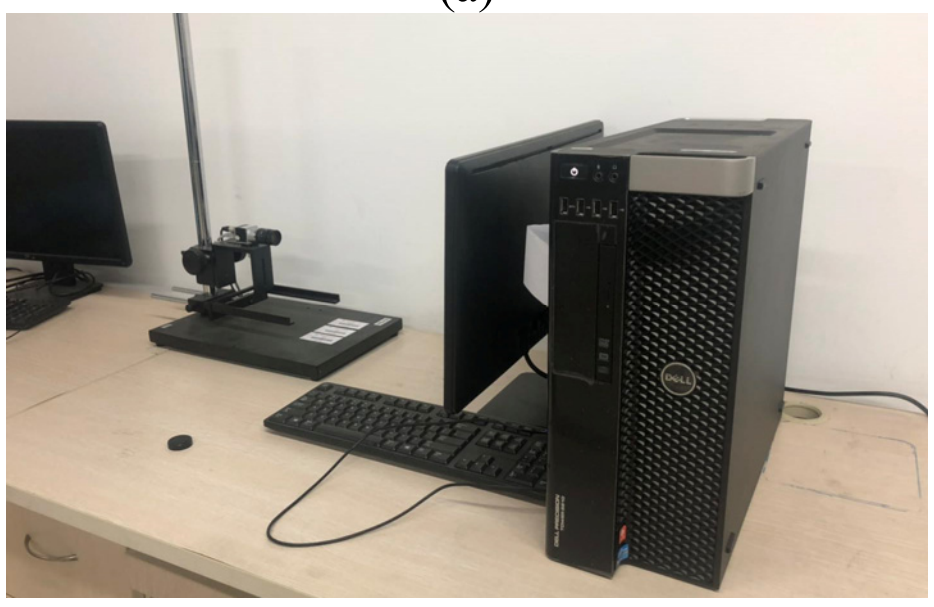

(b)

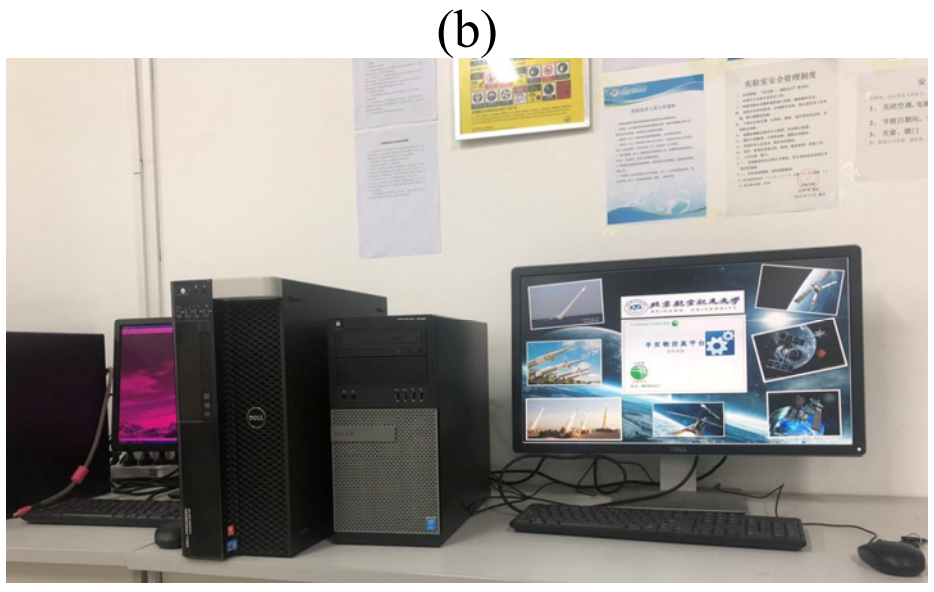

(c)

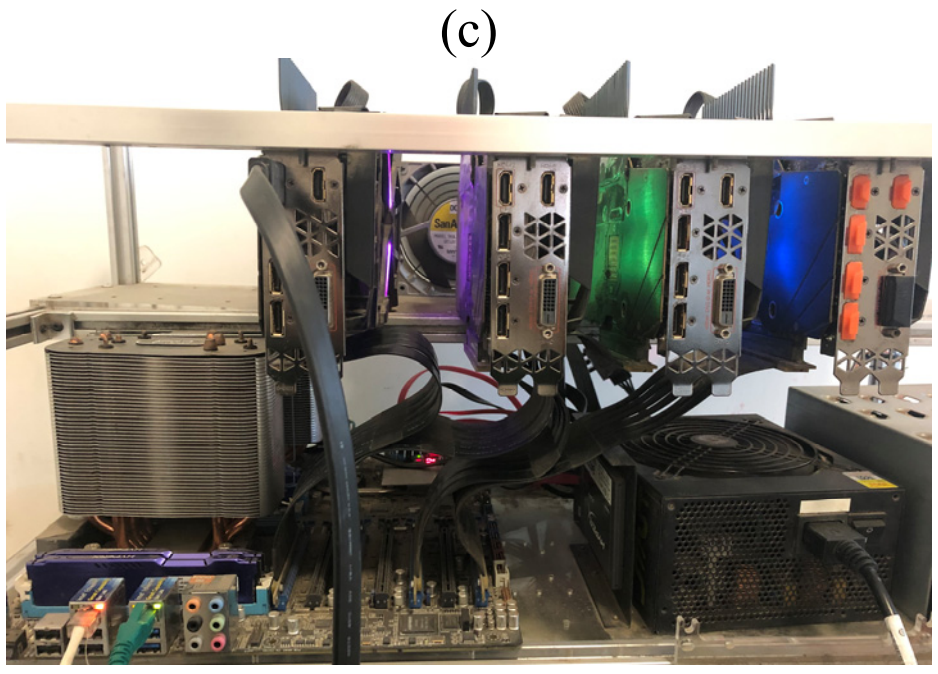

(d)

FIGURE 15. **Components of semi-physical simulation system. (a) Three-axis flight simulator. (b) Vision guidance subsystem. (c) Virtual reality subsystem (left) and main control computer (right). (d) The guidance computer.**

To verify the robustness of the proposed guidance law, experiments of 100 engagements are carried in the semi-physical simulation system. The desired LOS angles $\theta_{LD}$ and $\phi_{LD}$ are set to -0.8 rad and 0.6 rad respectively with the initial LOS angles set to a uniform distribution $Unif(-0.7,-0.9)$ and $Unif(0.5{,}0.7)$ respectively. The actuator gain $\eta$ is set to 0.6, which means it has a 40% loss of effectiveness. The targets have 30sin(t) m/s$^2$ acceleration in both directions. The $K_1, K_2$ in the cost function is set to $[0.9, 0.65]$ and $[8.0, 8.0]$ respectively, and the learning rate $\alpha$ for online adaptive control is reduced to 0.00065 to compensate the lag caused by missile autopilot. The observation $[R, \theta_m, \phi_m, \dot{R}, \dot{\theta}_m, \dot{\phi}_m]^T$ suffers from measurement uncertainty up to 8% in magnitude and measurement noise of LOS is chosen as Gaussian noise. Other initial engagement geometry parameters are same as the Monte Carlo case. The results are shown in Fig. 16 and Fig. 17, although lag caused by autopilot deteriorate online adaption and control performance, thus worsen interception performance compared with simulation results, experiment results still show positive terminal angle tracking performance and interception accuracy.

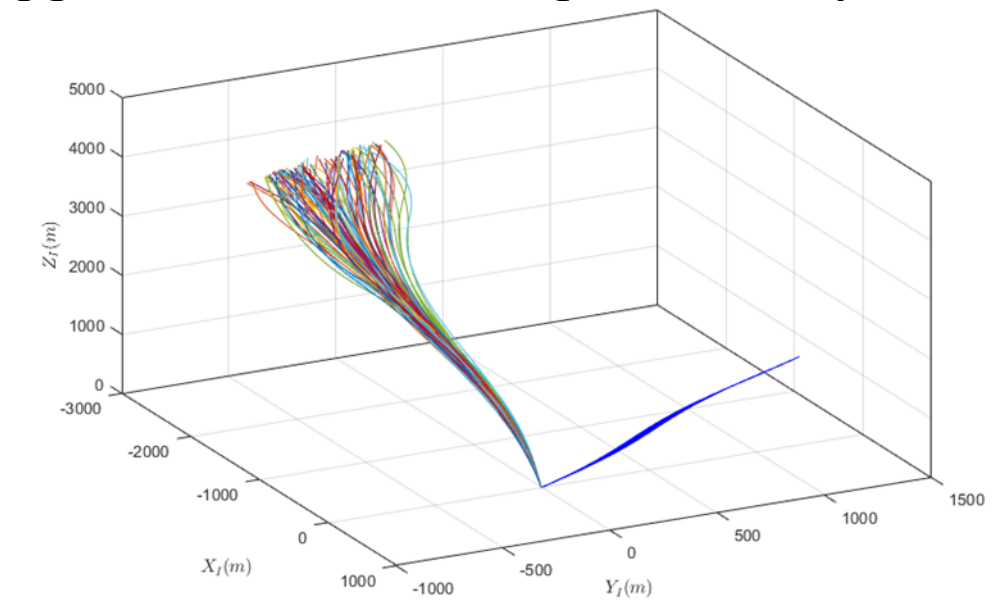

FIGURE 16. **Trajectories of interceptors (multi-color) and target (blue)**

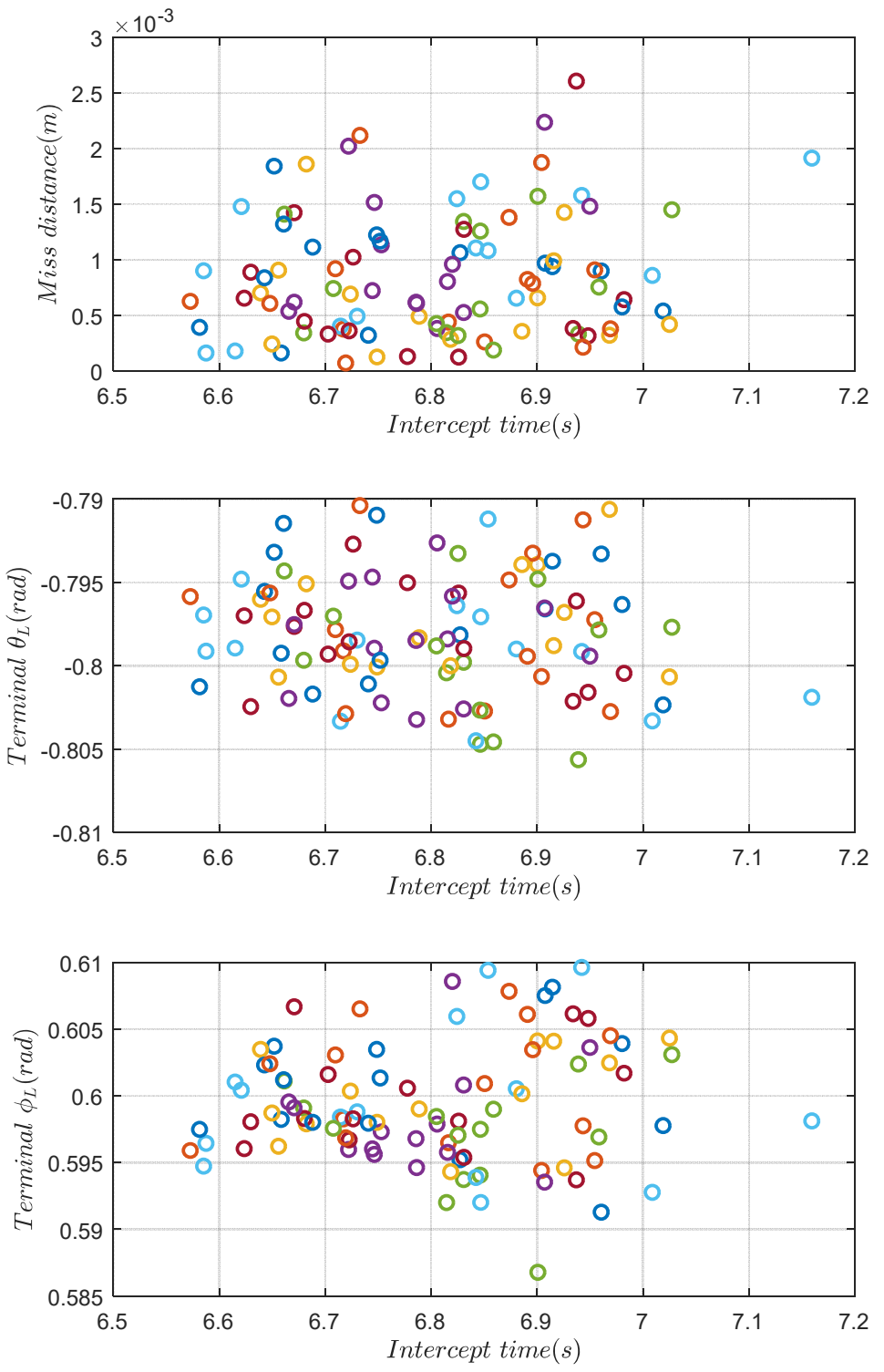

FIGURE 17. **Miss distance and Terminal LOS angles**

## V. CONCLUSION

In this paper, we present a novel impact-angle guidance law based on meta-learning and MPPI for the interception of a maneuvering target using a varying velocity interceptor under partial actuator failure. Model-based deep RL method is used in guidance law design and the guidance problem is solved with meta-learning MPPI control. In this framework a deep neural dynamic is used as a predictive model and the control command is computed using MPPI as a Monte Carlo sampled expectation over trajectories. With the online adaption ability provided by meta-learning, the deep neural dynamic in proposed guidance law can learn the change and perturbations in environment and thus has better tracking performance than standard MPPI method. Simulation and experiment under different cases are conducted to verify the performance and effectiveness of proposed guidance law in varying velocity interceptor intercepting maneuvering target under partial actuator failure.

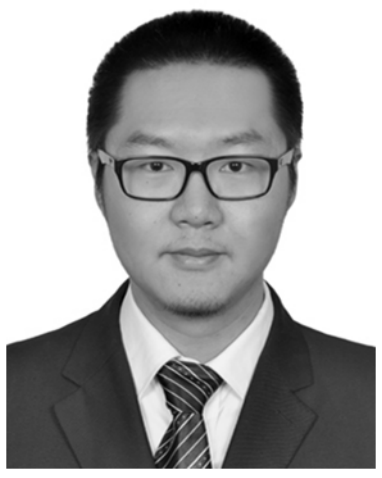

**CHEN LIANG** was born in Shandong, China, in 1990. He received the B.E. degree from Harbin Institute of Technology, Weihai, China, in 2012 and the M.S. degree in Electrical and Computer Engineering from Rutgers, the State University of New Jersey, New Brunswick, NJ, USA, in 2014. He is currently pursuing the Ph.D. degree in navigation, guidance and control at Beihang University, Beijing, China. His research interests include nonlinear control, machine vision, deep learning and guidance.

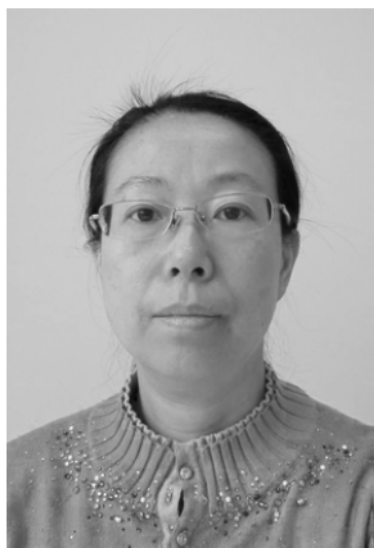

**WEIHONG WANG** received the B.E., M.S., and Ph.D. degrees from Harbin Institute of Technology, Harbin, China, in 1990, 1993, and 1996, respectively. She is currently a Professor with the School of Automation Science and Electrical Engineering at Beihang University, Beijing, China. Her research interests include computer simulation, computer control and simulation, guidance, intelligent control, servo control, and simulation technique.

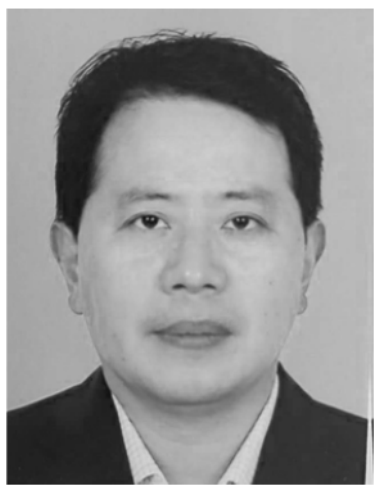

**ZHENGHUA LIU** received the B.E. and M.S. degrees from Nanjing University of Aeronautics and Astronautics, Nanjing, China, in 1997 and 2000, respectively, and the Ph.D. degree from the Beihang University, Beijing, China, in 2004. He is currently an associate professor with the School of Automation Science and Electrical Engineering, Beihang University, Beijing, China. His research interests include high precision servo control, robotic system, and flight control.

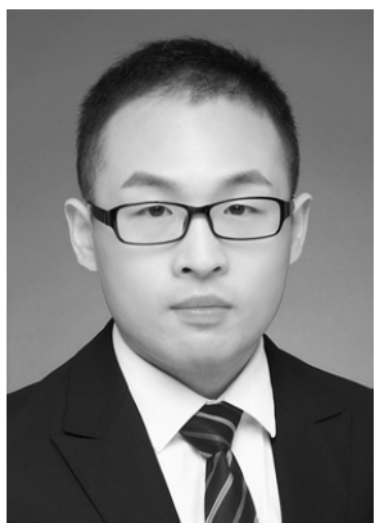

**CHAO LAI** was born in Shandong, China, in 1990. He received the B.E. degree from Dalian Maritime University and the Ph.D. degree from the Beihang University, Beijing, China, in 2019. He is currently with Navigation and Control Technology Research Institute of China North Industries Group Corporation (NORINCO). His current research interests include nonlinear control, servo control, adaptive control and integrated guidance and control.

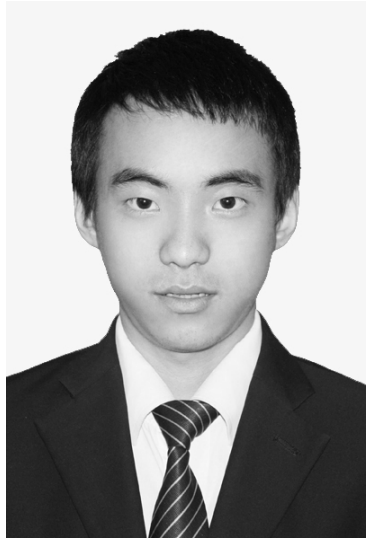

**BENCHUN ZHOU** received the B.E. degree from School of Automation, Chongqing University, Chongqing, China in 2012. He is currently pursuing the M.S. degree at Beihang University. His research interests include navigation, guidance and control of unmanned aerial vehicles, deep reinforcement learning and computer vision.